\pdfoutput=1

\documentclass[11pt]{article}

\usepackage{EMNLP2023}
\usepackage{graphicx}
\usepackage{times}
\usepackage{latexsym}

\usepackage{xcolor}         
\usepackage{colortbl}         
\usepackage{amssymb}

\usepackage{todonotes}

\usepackage[T1]{fontenc}

\usepackage[utf8]{inputenc}

\usepackage{microtype}

\usepackage{inconsolata}

\usepackage{forest}
\useforestlibrary{edges}

\usepackage{multirow}
\usepackage{booktabs}

\title{GenCodeSearchNet: A Benchmark Test Suite for Evaluating Generalization in Programming Language Understanding}

\author{Andor Diera \\
  \small{Ulm University,} \\
  \small{Germany} \\
  \small{\texttt{andor.diera@uni-ulm.de}} \And
  Abdelhalim Dahou \\
  \small{GESIS - Institute for Social Sciences,} \\  
  \small{Germany} \\  
  \small{\texttt{abdelhalim.dahou@gesis.org}} \\
  \\\And
  Lukas Galke \\
  \small{Max Planck Institute for Psycholinguistics,} \\
  \small{Netherlands} \\
 \small{\texttt{lukas.galke@mpi.nl}} \\
  \\\AND
  Fabian Karl \\
  \small{Ulm University,} \\
  \small{Germany} \\
  \small{\texttt{fabian.karl@uni-ulm.de}}
  \\\And
  Florian Sihler \\
  \small{Ulm University,} \\
  \small{Germany} \\
  \small{\texttt{florian.sihler@uni-ulm.de}} \\
  \And
  Ansgar Scherp \\
  \small{Ulm University,} \\
  \small{Germany} \\
\small{\texttt{ansgar.scherp@uni-ulm.de}} \\}

\begin{document}
\maketitle
\begin{abstract}
Language models can serve as a valuable tool for software developers to increase productivity. Large generative models can be used for code generation and code completion, while smaller encoder-only models are capable of performing code search tasks using natural language queries. 
These capabilities are heavily influenced by the quality and diversity of the available training data. Source code datasets used for training usually focus on the most popular languages and testing is mostly conducted on the same distributions, often overlooking low-resource programming languages. 
Motivated by the NLP generalization taxonomy proposed by Hupkes et.\,al., we propose a new benchmark dataset called GenCodeSearchNet (GeCS) which builds upon existing natural language code search datasets to systemically evaluate the programming language understanding generalization capabilities of language models. As part of the full dataset, we introduce a new, manually curated subset StatCodeSearch that focuses on R, a popular but so far underrepresented programming language that is often used by researchers outside the field of computer science.
For evaluation and comparison, we collect several baseline results using fine-tuned BERT-style models and GPT-style large language models in a zero-shot setting.
\end{abstract}

\section{Introduction}
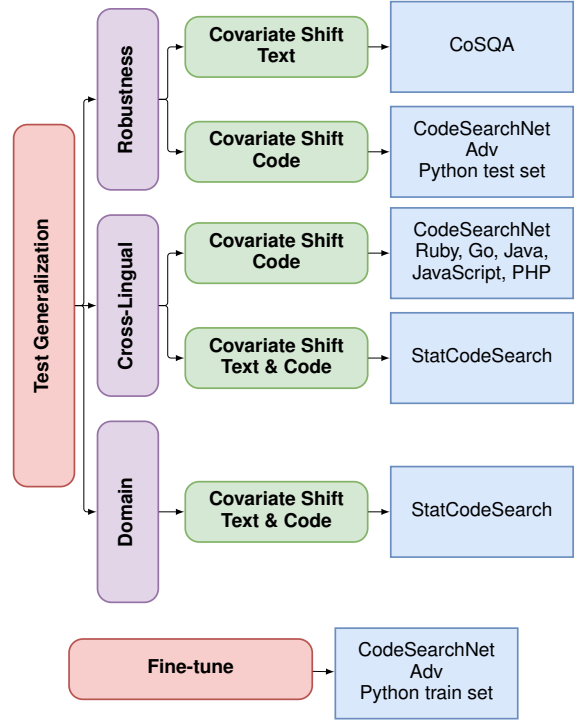
\begin{figure}[th]
    \definecolor{@0}{HTML}{B85450}%
    \definecolor{@0b}{HTML}{F8CECC}%
    \definecolor{@1}{HTML}{9673A6}%
    \definecolor{@1b}{HTML}{E1D5E7}%
    \definecolor{@2}{HTML}{82B366}%
    \definecolor{@2b}{HTML}{D5E8D4}%
    \definecolor{@3}{HTML}{6C8EBF}%
    \definecolor{@3b}{HTML}{DAE8FC}%
    \centering
    \forestset{d/.style={
        for tree={
            grow'=0,
            rectangle,
            rounded corners=6.5pt,
            draw,
            thick,
            anchor=west,
            fit=band,
            minimum width=3cm,
            minimum height=1cm,
            align=center,
            fork sep=4pt,
            edge={-latex,line cap=round,rounded corners=1.5pt},
            font={\sffamily\bfseries}
        },
        for root={
            minimum height=1cm,
            anchor=base,
            draw=@0,fill=@0b,minimum width=6cm,
            rotate=90
        },
        for level=1{draw=@1,fill=@1b,anchor=base,rotate=90},
        for level=2{draw=@2,fill=@2b},
        for leaves={draw=@3,fill=@3b,font=\sffamily\footnotesize,minimum height=1.5cm,sharp corners,,anchor=west,rotate=0}
    },last/.style={}}%
    \small
    \scalebox{.8}{\begin{forest}
        d
        [Test Generalization
            [Robustness,forked edge,
                [Covariate Shift\\Text,forked edge
                    [CoSQA\\]
                ]
                [Covariate Shift\\Code,forked edge,
                    [CodeSearchNet\\Adv\\Python test set]
                ]
            ]
            [Cross-Lingual,before computing xy={s/.average={s}{siblings}}
                [Covariate Shift\\Code,forked edge
                    [{CodeSearchNet\\Ruby, Go, Java,\\JavaScript, PHP}]
                ]
                [Covariate Shift\\Text \& Code,forked edge
                    [StatCodeSearch]
                ]
            ]
            [Domain,forked edge
                [Covariate Shift\\Text \& Code
                    [StatCodeSearch]
                ]
            ]
        ]
    \end{forest}}\\[2ex]
    \scalebox{.8}{\begin{forest}
        d
        [Fine-tune,rotate=0,minimum width=4cm
            [CodeSearchNet\\Adv\\Python train set,last]
        ]
    \end{forest}}
    \caption{Overview of the benchmark composition w.r.t. the generalization taxonomy of \citet{hupkes2022state}}
    \label{fig:framework}
\end{figure}

Language models have found their use in various tasks dealing with source code, ranging from code search to code summarization, code completion, and code translation~\cite{lu2021codexglue}. With the release of Codex~\cite{chen2021evaluating} and ChatGPT~\cite{ouyang2022training} large language models (LLMs) became popular and widely used for AI-assisted coding. Still, as of August 2023, code completion and code-related question answering with LLMs is far from reliable~\cite{kabir2023answers}. An alternative to using general purpose LLMs for code completion is code search (i.e., finding relevant source code based on a natural language query) in curated datasets~\cite{DBLP:conf/sigsoft/CambroneroLKS019,husain2019codesearchnet} which provides a more transparent aid for AI-assisted coding.
However, so far it is unclear how well code search models generalize across different programming languages, different domains, and how robust they are against distribution shifts.


Although both natural language queries and source code are represented as text, one cannot safely assume to have an overlap in the words/characters used, since function and variable names do not necessarily consist of words.  
Still, classical string-based matching between a query and documents~\cite{manning2009introduction} is used in many information retrieval systems and practical applications~\cite{lin2022pretrained}.
When documents consist of code, natural language queries will have little to no matching
words, resulting in a lexical gap. Due to this lexical gap, the task of finding
code snippets based on natural language queries is difficult and requires specific bimodal language models capable of processing both natural and programming languages~\citep{feng2020codebert,guo2020graphcodebert,wang2021codet5,wang2023codet5+}.

However, this bimodal training approach is likely limited to the programming languages on
which the models were trained. It is unknown how such models would generalize to
programming languages that were not part of the training data or have only little representation in the dataset, i.\,e., a low-resource programming language. 
Evaluating the models on programming languages that were not part of the training data (i.\,e., a distribution shift occurs) would shed new light
on the generalization capabilities of hybrid models for code and text -- which
is the aim of this work.


Moreover, there are low-resource programming languages such as R that in terms
of quantity are underrepresented on popular code-sharing platforms like GitHub.
However, it is the de~facto programming language in many research fields relying on statistical analysis, such as economics, statistics, social sciences, and psychology. Thus, the coding conventions and style in these fields also differ from the code corpora used in existing benchmark datasets~\cite{husain2019codesearchnet, lu2021codexglue}.
This produces a blind spot on the current methods for code search as they are usually only tested on datasets of well-curated source code in the most popular programming languages.

We propose a new benchmark dataset called GenCodeSearchNet (GeCS) which combines  
a new, manually curated dataset StatCodeSearch with existing code search datasets. StatCodeSearch consists of code-comment pairs extracted from R scripts written for statistical analysis.
We further propose an evaluation protocol for the benchmark that allows future researchers to systemically test the language models' generalization capabilities for programming language understanding. 
The evaluation setup for our dataset is illustrated in Figure~\ref{fig:framework} and consists of three generalization tests for robustness, cross-lingual, and domain generalization.
We provide a detailed description of this benchmark in Section~\ref{sec:composition}.
In summary, the contributions of this work are three-fold:
\begin{itemize}
    \item We create a benchmark for \emph{programming language understanding} named GenCodeSearchNet that tests text-code matching and ranking, organized along different types of out-of-distribution generalization. The composition of the benchmark is described in Section~\ref{sec:composition} and its evaluation protocol in Section~\ref{sec:eval}.
    \item To facilitate the new benchmark, we introduce a new, manually-curated dataset named StatCodeSearch, consisting of 1,070 text-code pairs from statistical research code written in R, which is described in Section~\ref{sub:new-dataset}.
    \item Initial baselines for this new benchmark are introduced in Section~\ref{sec:baselines}.
    We provide results for RoBERTA, CodeBERT, CodeT5+, and GPT-based LLMs in Section~\ref{sec:results}.
\end{itemize}



\section{Related Work}
\label{sec:relatedwork}

\paragraph{Code Search} Code search is an established research field with various tools and solutions available.
Below, we briefly summarize existing classical works on source code search, followed by works on semantic code search based on neural networks, particularly pre-trained language models.

An established classical tool for source code search is Oracle OpenGrok\footnote{\url{https://github.com/oracle/opengrok}} based on the popular full-text index Lucene. As a result, OpenGrok enables textual searching of code for strings based on Google-like search queries. Similar systems for textual searching on code are searchcode\footnote{\url{https://searchcode.com/ }} and Sourcegraph\footnote{\url{https://about.sourcegraph.com/}}.
The ANNE~\cite{VSP17} system extends the purely textual search to source code by mapping natural language queries to syntactic keywords of programming languages. 
Search engines supporting structured queries include Aroma~\cite{LYB19}, which takes an incomplete code fragment as a query (called a snippet) and suggests concise code snippets from the code database. 
A similar approach is also taken by Mukherjee et al.~\cite{MJC20}. 

Neural networks have also been successfully applied for code search allowing semantics-based natural language code search.
Just as in most fields of NLP, the best-performing models for semantic code search are transformer-based language models. These models are pretrained on both natural language and programming language corpora and often use a contrastive loss to better align text and code representations~\cite{li2022coderetriever,wang2023codet5+,neelakantan2022text}. Prominent examples of encoder-only bimodal language models include CodeBERT~\cite{feng2020codebert}, GraphCodeBERT~\cite{guo2020graphcodebert}, and CodeRetriever~\cite{li-etal-2022-coderetriever}.
Encoder-decoder models designed to handle a wide range of programming language tasks such as CodeT5~\cite{wang2021codet5} and CodeT5+~\cite{wang2023codet5+} also perform strongly on the task of semantic code search. Lastly, decoder-only models~\cite{brown2020language} can be also employed for code search, albeit they either require careful prompting~\cite{ouyang2022training} or extensive fine-tuning~\cite{neelakantan2022text}.

\paragraph{Generalization in Programming Language Understanding}
Generalization, or the ability of a model to perform well on data not seen during training, is sought after in all domains of machine learning~\cite{goodfellowDeepLearning2016}. 
However, generalization can refer to a wide range of different scenarios in NLP.
To tackle this lack of agreement and systematic testing, \citet{hupkes2022state} proposed a taxonomy for characterizing generalization research in NLP. Their taxonomy consists of five axes to classify the motivation, generalization type, data shift type, source, and locus of the shift.

Even though neural networks capable of handling both natural and programming languages are usually called bimodal models~\cite{allamanis2015bimodal,feng2020codebert,wang2023codet5+}, the representations of these two input modalities share a lot of commonalities, including predictable statistical properties~\cite{naturalness_of_sw_hindle}. 
Therefore, we argue that the NLP generalization taxonomy proposed by~\citet{hupkes2022state} also offers valuable insights for research in programming language understanding.

Generalization research on code-related tasks is still in short supply. The HumanEval dataset~\cite{chen2021evaluating} used for benchmarking generative models only contains Python source code, while CodeXGLUE~\cite{lu2021codexglue}, the comprehensive programming language understanding benchmark suite is designed to evaluate on in-distribution test data. CodeS~\cite{hu2023codes} offers an extensive dataset for evaluating against different types of out-of-distribution samples, but only uses two languages (Python and Java) with the singular downstream task of code classification. 
To the best of our knowledge, the only large-scale benchmarks for evaluating generalization of code-related tasks are CrossCodeBench~\cite{niu2023crosscodebench} and XLCoST~\cite{zhu2022xlcost}. 
While XLCoST focuses solely on cross-lingual generalization, CrossCodeBench only includes tasks that are formulated in a text-to-text form, leaving out retrieval tasks, such as code search. In contrast to the aforementioned works, our proposed benchmark focuses on the task of natural language code search and offers evaluations against multiple types of distribution shifts.

\section{Composition of GenCodeSearchNet}\label{sec:composition}

\begin{table*}[ht]
    \centering
    \small
    \caption{Statistics of the different subsets used in the GenCodeSearchNet test suite. The numbers shown include only the positive (matching) examples}
    \begin{tabular}{l|r|r|r|r|r}
    \hline
    \textbf{Subset} & 
       \textbf{\# text-code} & 
       \textbf{avg \# text} & 
       \textbf{avg \# code} & 
       \textbf{total variation} & 
       \textbf{total variation} \\
    \textbf{ } & 
       \textbf{pairs} & 
       \textbf{tokens} & 
       \textbf{tokens} & 
       \textbf{distance text} & 
       \textbf{distance code} \\
    \hline
    \textbf{Fine-tuning set} & & & & & \\
    CodeSearchNet AdvTest train & 251\,820 & 15.97 & 166.94 & 0.0 & 0.0\\
    \hline
    \textbf{Test sets} & & & & & \\
    CodeSearchNet AdvTest test& 19\,210 & 16.08 & 177.64 & 0.1268 & 0.5372\\
    CodeSearchNet Go & 14\,291 & 27.46 & 159.08 & 0.3281 & 0.6110\\
    CodeSearchNet Java & 26\,909 & 29.07 & 179.63 & 0.3126 & 0.5972\\
    CodeSearchNet JavaScript & 6\,483 & 21.93 & 240.21 & 0.2892 & 0.5479\\
    CodeSearchNet Ruby & 2\,279 & 23.88 & 138.09 & 0.2962 & 0.5610\\
    CodeSearchNet PHP & 29\,391& 14.73 & 189.00 & 0.2858 & 0.5819\\
    CoSQA & 10\,293 & 10.42 & 55.93 & 0.5322 & 0.4453\\
    StatCodeSearch & 1\,070 & 24.55 & 134.02 & 0.5386 & 0.8032\\
    \hline
    \textbf{Combined Test set}& 109\,926 & 21.01 & 159.20 & 0.3386 & 0.5855\\
    \hline
    \end{tabular}
    \label{tab:dataset_stats}
\end{table*}

In this section, we describe the datasets used in our benchmark suite. The datasets are chosen and created based on the criterion to evaluate different types of generalization in programming language processing to foster further research in the field.

The GeCS dataset includes one fine-tuning set and eight test sets. It contains three previously proposed datasets, namely CodeSearchNet~\citep{husain2019codesearchnet}, CodeSearchNet AdvTest~\citep{lu2021codexglue}, and CoSQA~\citep{huang2021cosqa}. 
We add a novel set that focuses on statistical tests in the programming language R, named StatCodeSearch. Each dataset contains a natural language description (either a code comment or 
a search engine query) and a source code snippet. The test suite is designed to study generalization from a practical perspective, i.\,e., to assess programming language understanding in various evaluation scenarios.
The source of datashifts between the different test sets is considered to be naturally occurring, with the locus of the shift appearing between the pretraining/fine-tuning data and test data. The types of generalization covered by the test suite include robustness to covariate shifts and generalization across programming languages and programming domains.

A detailed breakdown of the experimental design in the generalization framework proposed by~\citet{hupkes2022state} can be seen in Figure~\ref{fig:framework}.
The main characteristics of the different subsets can be found in Table~\ref{tab:dataset_stats}. The average number of tokens have been calculated using the pretrained RoBERTa tokenizer sourced from HuggingFace~\cite{wolf2019huggingface}. Furthermore, we also measure the covariate shift in texts and codes using the total variation distance~\cite{goldenberg2019survey}. We calculate the total variation distance to the CodeSearchNet Adv train set (used for fine-tuning) on the tokenized samples.

\subsection{Existing Datasets}

\label{sec:codesearchnet}
The \textbf{CodeSearchNet} dataset~\cite{husain2019codesearchnet} was introduced as a semantic code search evaluation tool.
Since then it has been a staple benchmark dataset for studying the code search capabilities of machine learning models. 
It encompasses code-comment pairs from six different programming languages: Python, Go, Java, JavaScript, Ruby, and PHP. The full corpus includes 6 million functions scraped from GitHub, with 2 million of those including associated function documentation. Functions less than three lines and documentation shorter than three tokens were removed from the scraped corpus. Duplicate or near duplicate functions were also discarded to control for auto-generated code snippets and copy~\& paste between GitHub users. For the GeCS test suite, we collected the test sets from the HuggingFace Hub~\cite{lhoest2021datasets} and 
discarded the Python subset (since it is included later in the CodeSearchNet AdvTest dataset). 
We employed no further preprocessing and formatted the data into JSONL files. 
For each sample, we defined three fields: the \textit{input} field contains the code comment and code snippet separated by a '[CODESPLIT]' token, the \textit{target} field contains the index of the binary labels ('no\_match','match'), which are found in the \textit{target\_options} field. This dataset is used to measure cross-lingual generalization.

The \textbf{CodeSearchNet AdvTest} dataset was developed for the CodeXGLUE benchmark dataset~\cite{lu2021codexglue} by applying further preprocessing steps on the Python subset of the CodeSearchNet dataset. 
First, all code snippets that could not be parsed into an abstract syntax tree were removed, then documentations with more than~256 tokens were removed alongside samples that contained special tokens such as \textit{"http://"} or \textit{"<img... >"}. Finally, the functions and variables in the code snippets of the test set have been normalized by renaming them to \textit{func} and \textit{arg\textsubscript{i}}, respectively. This normalization of the test set makes it a good fit to test robustness against a covariate shift in the source code. For the inclusion in the GeCS test suite, we sourced both the train and test set from the official CodeXGLUE repository\footnote{https://github.com/microsoft/CodeXGLUE}.
We employed no further preprocessing and applied the same formatting as described above.

The Code Search And Question Answering~(\textbf{CoSQA}) corpus~\cite{huang2021cosqa} uses the Python subset of the CodeSearchNet dataset and matches the code snippets with real-world search queries from the Microsoft Bing search engine. The search logs were carefully filtered to only include queries that incorporate the keyword \textit{python} and have none of the predefined keywords that relate to search intents other than code search (e.\,g.,  debugging, conceptual queries, tool usage). After this initial rule-based filtering, a fine-tuned CodeBERT encoder~\cite{feng2020codebert} was used to measure cosine similarity between candidate queries and code snippets. Each code snippet was then matched with the query of the highest similarity. 
Pairs with a similarity value of less than 0.5 were discarded. 
Finally, a number of human annotators were instructed to label each code-query pair whether the code snippet answers the query or not. 
For the GeCS dataset, we use the training set retrieved from the official CodeXGLUE repository and discard query-code pairs that are labeled as non-matching (we create our own negative samples as described in Section~\ref{sec:eval}). We apply no further preprocessing and use the same formatting as described before.

\subsection{New Dataset: StatCodeSearch}
\label{sub:new-dataset}
The StatCodeSearch dataset is a benchmark test set consisting of code comment pairs extracted from R programming language scripts authored mostly by researchers. 
The dataset is sourced from the Open Science Framework (OSF)\footnote{https://osf.io/}. It includes text and code samples from R projects that pertain to the fields of social science and psychology with a focus on the statistical analysis of research data. These projects are often linked with research articles published in journals such as \textit{Political Communication}, \textit{Behavior Research Methods}, and \textit{Cognitive Science}.
R~scripts in these domains seldom use branching or explicit looping constructs, as most logic is handled by higher-order functions and R's implicit vectorization. 
Furthermore, they heavily rely on functions of loaded libraries~\cite{sihler2023slicer}.

The initial scraping of the OSF website resulted in 11,775 R projects. After discarding projects that did not have any specific permissive software license, the dataset was narrowed down to 2,832 projects, from which the code-comment pairs of the final dataset were extracted.
The creation of code-comment pairs involved a three-step procedure. First, we implemented a rule-based extraction, which included the following steps. 
We removed empty lines and leading spaces from each line.
We discarded lines identified as library loading.
We detected lines commencing with \textit{"\#"} that include more than one word. 
If multiple subsequent comment lines were found, we concatenated them into one text item.
We categorized lines without the leading \textit{"\#"} symbol as associated code blocks to the preceding text item. 

After the extraction of code-comment pairs through these rules, an additional post-processing step was applied where we discarded common comment trailing symbols from the text items such as \textit{"\#"}, \textit{"-"}, and \textit{"="}. This first step resulted in~40,041 pairs of code and comments. In order to filter out irrelevant comments, in the second step we employed GPT~3.5 Turbo to classify the comments into four predefined classes (the prompts used for this subtask can be found in Appendix~\ref{sec:prompts}). These classes were defined as statistical tests, statistical modeling, visualization, and data variables.
On a small subset of 400 pairs, we experimented with three prompting methods for this task: zero-shot, one-shot, and few-shot. Our pre-experiments showed zero-shot to be the most suitable approach for this filtering, as it produced fewer false-negatives than the other two approaches. This automated filtering resulted in a total of~10,137 code comment pairs.
The third step involved the review and evaluation of the remaining selection by the authors. 
We manually filtered the remaining 10,137 code pairs and removed those with irrelevant comments, or code blocks that did not correspond to the comment. This step served as a critical quality control measure and yielded a total of 1,070 pairs of code and comments. We applied the same formatting as described in Section~\ref{sec:codesearchnet}.

\section{Evaluation Protocol}
\label{sec:eval}

%
%
%

\subsection{Fine-tuning}
Fine-tuning is an optional step for evaluating smaller models that did not have a text-code matching objective during their pretraining phase. For a fair comparison, we suggest only using the training set of the CodeSearchNet AdvTest dataset for fine-tuning, which is based on Python, a widely-used general-purpose programming language.

\subsection{Measures}\label{sub:measures}

We apply two measures for assessing the performance of the models.

\paragraph{Matching}
We test whether a given text-code pair is a matching pair (positive) or not (negative).
We evaluate the accuracy on balanced test sets with an equal number of positive and negative examples. 
The non-matching examples are sampled uniformly within the respective dataset.

\paragraph{Ranking}
To evaluate the ranking in the code search task, we employ Mean Reciprocal Rank~(MRR). For each query, we consider 99 distractors sampled uniformly at random.
For each query, the reciprocal of the best-ranked correct answer is considered.
Formally, for a set of queries~$Q$, the reciprocal of the best-ranked correct answer at rank~$r_i$ is aggregated and averaged as $\mathrm{MRR} = \frac{1}{|Q|} \sum_{i=1,\ldots, |Q|} \frac{1}{r_i}\,$.


The rationale for the choice of MRR over 
alternative ranking metrics, such as mean average precision (MAP) or normalized discounted cumulative gain (nDCG)~\cite{manning2009introduction,lin2022pretrained}, 
is that we have only a single relevant code snippet for each query and the ratings are binary.
When there is only a single relevant document, as in CodeSearchNet, MRR and MAP coincide to the same formula.
Furthermore, nDCG can reflect different degrees of relevance, but in our case, since we only have a binary assessment of the code-comment pairs, there is no benefit of using this metric.

\subsection{Evaluation by Generalization Type}
\label{gen_type}
Our proposed benchmark groups the evaluations by the generalization types proposed by~\citet{hupkes2022state}. For this, we take the unweighted average of the classification and ranking scores across different datasets (see Figure~\ref{fig:framework}).
%
To evaluate \textbf{robustness}, we aggregate scores on test sets that exhibit a covariate shift in either text or code. For this, we employ CoSQA and CodeSearchNet AdvTest (as described in Section~\ref{sec:codesearchnet}).
%
For \textbf{cross-lingual generalization} (cross-lingual referring to ``across programming languages''), we average the scores across datasets CodeSearchNet and StatCodeSearch. In this generalization type, the locus of the covariate shift is mainly in the code snippets due to the differing syntax.
%
For \textbf{domain generalization}, the sole test set is StatCodeSearch from the domain of statistical analysis, which entails a different coding/commenting style (cf. Section~\ref{sub:new-dataset}). 


\section{Baselines}
\label{sec:baselines}

\subsection{Baseline Methods}

We provide the results for three main types of baseline models using two evaluation strategies. We employ the encoder-only models RoBERTa and CodeBERT, the encoder-decoder model CodeT5+, and GPT-based models GPT 3.5 Turbo and Text-embedding-ada-002. For RoBERTa and CodeBERT we employ an additional fine-tuning step on the CodeSearchNet AdvTest train set. The GPT-based models are evaluated in a zero-shot setup, while CodeT5+ is tested both in the fine-tuning and zero-shot setup.

For fine-tuning, we follow the same sampling procedure that is also in the evaluation of the matching task, i.\,e., single negative example for matching. 
To tackle the matching task, we concatenate the query and code to a singular input and train an output layer on binary classification.
To tackle the ranking task with fine-tuned models, we concatenate the query with each of the 100 candidate code snippets. Then, we rank all 100 candidate pairs according to the matching score emitted by the model. In the zero-shot ranking setups we create the ranking based on the cosine similarity between text and code inputs. The choice of hyperparameters for fine-tuning can be found in Appendix~\ref{sec:hyperparams}.
Below we briefly describe the existing models that we use as initial baselines for the proposed benchmark.  

\paragraph{RoBERTa} is an encoder-only transformer model that builds upon the foundations of the BERT model~\cite{devlin2019bert}. RoBERTa is designed to improve upon some limitations of BERT by optimizing the pretraining process. These improvements include dynamic masking in masked language modeling, larger batch sizes, increased training data size, and a more thorough hyperparameter optimization~\cite{liu2019roberta}. RoBERTa is a commonly used baseline model for numerous NLP tasks, including semantic code search~\cite{feng2020codebert}. Our experiments are based on the pretrained HuggingFace implementation of the \textit{RoBERTa-base} model ~\cite{wolf2019huggingface}.

\paragraph{CodeBERT} is a pretrained transformer model designed for both natural language and programming language tasks~\cite{feng2020codebert}. It uses the RoBERTa-base architecture with a masked language modeling objective for bimodal (natural language and programming language input pairs) training data and replaced token detection for unimodal (only programming language input) training data. Similarly to other BERT-based models, CodeBERT performs best with task-specific fine-tuning, and is a common baseline in programming language understanding tasks such as semantic code search, code summarization, and code-clone detection~\cite{lu2021codexglue}. We use the pre-trained CodeBERT model from HuggingFace~\cite{wolf2019huggingface} and follow the original paper's fine-tuning procedure.

\paragraph{CodeT5+} is an encoder-decoder transformer model suited to solve a wide range of code tasks~\cite{wang2023codet5+}. This is made possible by employing a mixture of pretraining objectives, including span denoising, contrastive learning, text-code matching, and causal language modeling on both unimodal and bimodal training data. 
The CodeT5+ model can be be successfully deployed in multiple settings (zero-shot, fine-tuning, and instruction-tuning) and performs very well on over 20 code-related benchmark tasks. We run our experiments with the \textit{codet5p-110m-embedding} model variant sourced from HuggingFace.
This version, denoted as \emph{CodeT5+ (encoder only)} in our experiments, includes only the encoder layers of the bimodal model. This makes it suitable to create high quality embeddings with lower computational costs. For the matching evaluation, we extend this model with a binary classifier head (similar to the RoBERTa and CodeBERT setups) and apply fine-tuning. In the ranking evaluation, we use both the fine-tuned version and the base model in a zero-shot setup. 

\paragraph{GPT-3.5 Turbo} developed by OpenAI as a member of the GPT family~\cite{brown2020language} is specifically designed to understand and generate both natural language and code.  GPT-3.5 Turbo is optimized primarily for chat applications but also excels in traditional understanding and completion tasks. 
The model is also a successor to the Open\-AI Codex model~\cite{chen2021evaluating} and has exhibited significant enhancements in code generation, error detection, debugging, and analysis.
Like other LLMs, it is built on the transformer architecture, is pre-trained on vast amounts of text and code, and is heavily fine-tuned through human feedback~\cite{ouyang2022training}.
We employ GPT-3.5 for the binary classification evaluation in a zero shot setup (prompts can be found in Appendix~\ref{sec:prompts}).

\paragraph{Text-embedding-ada-002} denoted as Ada 2 in our experiments, is an embedding model released by OpenAI in late 2022. Embedding models are designed to generate vectors of floating point numbers that capture the semantic meaning of the input~\cite{neelakantan2022text}. The second generation of the OpenAI Ada model has been created by merging the functionalities of five distinct embedding models related to text search, text similarity, and code search into one unified interface. 
We employ Text-embedding-ada-002 to calculate the cosine similarity between the embeddings of the input pairs in the ranking evaluation setup.

\subsection{Baseline Results}\label{sec:results}
The evaluation results for the GeCS dataset are shown in Table~\ref{tab:all_results}. The aggregated results for each generalization type (as described in Section~\ref{gen_type}) are displayed in Table~\ref{tab:gen_type_results}.
A breakdown of results by programming language within CodeSearchNet can be found in Appendix~\ref{app:breakdown}.

\paragraph{Matching}
The fine-tuned models achieve on average around 90\% accuracy on the matching task, with RoBERTa producing both the lowest (84.41\% on CodeSearchNet AdvTest) and highest (99.18\% on CodeSearchNet Go) results. The matching results of GPT~3.5 Turbo range from 32.82\% (CoSQA) to 62.71\% (StatCodeSearch). 
The highest results across the fine-tuned models were achieved on the CoSQA and CodeSearchNet Go datasets, while the lowest values were seen in the PHP subset of CodeSearchNet. 
Aggregating by generalization type, we find that CodeT5+ yields the highest scores on Robustness, while CodeBERT yields the highest scores on Cross-Lingual and Domain.

\paragraph{Ranking}
In the ranking evaluation, the highest MRR ratings are achieved by the Ada 2 model, which consistently places the correct code snippet in the first two ranks on average. The zero-shot CodeT5+ models also attain similarly strong performance. Compared to the zero-shot models, fine-tuned models performed poorly on ranking, achieving the highest MRR scores on the robustness tests (CodeSearchNet AdvTest and CoSQA), while scoring below 0.1 on the cross-lingual and domain tests. The highest results across all models were obtained on the CodeSearchNet Ruby datasets, while the lowest values are seen in the StatCodeSearch dataset.
Aggregating by generalization type, we find that all models yield higher scores for Robustness than for Cross-Lingual and Domain.


\begin{table*}[ht]
    \small
    \centering
    \caption{Baseline Results on GenCodeSearchNet}
    \begin{tabular}{l|rr|rr|rr|rr}
    \hline
    \textbf{Model} & \multicolumn{2}{c}{\textbf{CodeSearchNet Avg}} & \multicolumn{2}{c}{\textbf{CodeSearchNet AdvTest}} & \multicolumn{2}{c}{\textbf{CoSQA}} & \multicolumn{2}{c}{\textbf{StatCodeSearch}} \\
     & Acc & MRR & Acc & MRR & Acc & MRR & Acc & MRR \\
    \hline
    \textbf{Fine-tuned Models} & & & & & & \\
    RoBERTa & 0.9263 & 0.1054 & 0.8441 & 0.3853 & 0.9596 & 0.0441 & 0.8958 & 0.0557\\
    CodeBERT & 0.9056 & 0.0907 & 0.8862 & 0.4191 & 0.9758 & 0.1087 & 0.9607 & 0.0251\\
    CodeT5+ (encoder only) & 0.8734 & 0.0616 & 0.9002 & 0.2767 & 0.9773 & 0.0482 & 0.9056 & 0.0582\\
    \hline
    \textbf{Zero-shot Models} & & & & & & \\
    CodeT5+ (encoder only) & - & 0.8198 & - & 0.8547 & - &  0.7972 & - & 0.6311 \\
    GPT 3.5 Turbo & 0.5882& - & 0.5687 & - &0.3282 & - & 0.6271 & -\\
    Ada 2 & - & 0.8852 &- & 0.8264 &- &0.9439 & -&  0.7945\\
    \hline
    \end{tabular}
    \label{tab:all_results}
\end{table*}

\begin{table*}[ht!]
    \small
    \caption{Aggregated Performance for Each Generalization Type}
    \centering
    \begin{tabular}{l|rr|rr|rr|rr}
    \hline
    \textbf{Model} & \multicolumn{2}{c}{\textbf{Robustness}} & \multicolumn{2}{c}{\textbf{Cross-Lingual}} & \multicolumn{2}{c}{\textbf{Domain}} & \multicolumn{2}{c}{\textbf{Combined}} \\
    & Acc & MRR & Acc & MRR & Acc & MRR & Acc & MRR \\
    \hline
    RoBERTa & 0.9018 & 0.2147 & 0.9110 & 0.0322 & 0.8958 & 0.0557 & 0.9028 & 0.1008 \\
    CodeBERT & 0.9310 & 0.2639 & 0.9170 & 0.0579 & 0.9607 & 0.0251 & 0.9362 & 0.1236\\
    CodeT5+ (encoder only) FT & 0.9387 & 0.1156 & 0.8895 & 0.0599 & 0.9056 & 0.0582 & 0.9112 & 0.0779 \\
    CodeT5+ (encoder only) ZS & - & 0.8259 & - & 0.7254 & - & 0.6311 & - & 0.7274\\
    GPT 3.5 Turbo & 0.4485 & - & 0.6076 & - & 0.6271 & - & 0.5610 & -\\
    Ada 2 & - & 0.8851 & - & 0.8398 & & 0.7945 & - & 0.8398\\
    \hline
    \end{tabular}
    \label{tab:gen_type_results}
\end{table*}

\section{Discussion}
Our newly introduced benchmark dataset provides several new insights on the generalization capabilities of pre-trained language models. Compared to existing benchmarks, such as CodeSearchNet and CodexGLUE, it focuses on evaluating different types of out-of-distribution generalization, leading to new insights about existing models.

First, we observe that encoder-only models completely fail at ranking when tested out of distribution
(on datasets for which they were not specifically fine-tuned).
%
In general, we find that ranking performance suffers substantially from fine-tuning, suggesting that the models are overfitting to the fine-tuning set, resulting in limited generalization capabilities in all three investigated
generalization types.

On the other hand, large-scale pre-trained models, especially Ada 2, excel at ranking on all datasets. 
The dataset on which such zero-shot models yield the lowest performance is the newly introduced
StatCodeSearch. 
A possible explanation is that the other 
datasets originate from GitHub and likely suffer from contamination, i.e., their test data could be part of the training data of the large language models~\cite{golchin2023time}.  
%

The low performance (hardly better than chance) of GPT-3.5 Turbo in zero-shot matching is surprising. 
We cannot exclude that the performance could be increased by providing a few examples in the prompt. We opted for testing its zero-shot capabilities for a fair comparison with the other LLMs, leaving room for future work with more refined prompting strategies.

%

Reflecting \citet{hupkes2022state}'s generalization framework for code-related tasks has allowed us to better understand how the generalization type affects language model performance  
on tasks involving both natural and programming language understanding:
Overall our results suggest that fine-tuned encoder-only models are strong at matching even in out-of-distribution test sets, while large-scale embedding models are strong at zero-shot ranking.
Bimodal language models have been shown to achieve high performances on in-distribution natural language code search~\cite{feng2020codebert,wang2023codet5+}. 
Their results on our benchmark indicate shortcomings of existing models when tested against out-of-distribution data.
We hope our newly introduced benchmark can facilitate the development of bimodal language models that generalize well beyond the training or fine-tuning distribution.

\section{Conclusion} We have introduced a new benchmark called GenCodeSearchNet (GeCS)
for testing the generalization capabilities of language models. The
benchmark specifically aims at scrutinizing different types of generalization
(robustness, cross-lingual, and domain), and evaluates matching and ranking for
each type. 
To test generalization with covariate shifts in both textual descriptions and code snippets, we provide a new, manually curated dataset StatCodeSearch with
R code comment pairs harvested from OSF. 
As initial baselines, we have evaluated RoBERTa, CodeBERT, CodeT5+, GPT 3.5 Turbo, and Ada 2. 
The baseline results of our benchmark
reveal that models that are good at matching are not necessarily good at
ranking and vice-versa.  
Hence, we hope that the new benchmark spurs the development of programming language-agnostic models that are good at both ranking and matching.
%
%
Moving forward, one could extend GeCS with other low-resource programming languages to further facilitate the systematic evaluation of pre-trained language models against different aspects of generalization.  
We make our experiments and
baseline models available on GitHub\footnote{\url{https://github.com/drndr/gencodesearchnet}}. The final dataset is available on Zenodo\footnote{\url{https://doi.org/10.5281/zenodo.8310891}}.

\section*{Acknowledgements}
This research is co-funded by the CodeInspector project (No. 504226141) of the DFG, German Research Foundation. 
The authors also acknowledge support by the state of Baden-Württemberg through bwHPC computing infrastructure. We would also like to thank Saurav Karmakar, Marcel Hoffmann, and Nicolas Lell for their insightful comments and reviews on our work.

\section*{Limitations}
As with most benchmark collections, there is a risk that part of the test data has
been present in the pre-training data of a large language model. As the new
dataset StatCodeSeach is not based on code harvested from GitHub but from OSF, we assume that current language models did not have access to it in their pre-training data. 
However, due to the intransparency of the training set of
corporate language models, we cannot fully exclude it. 

We did not apply Ada 2 to the matching task. Although such embedding models can be also adapted for matching tasks, doing so in a zero-shot fashion is not straightforward as it would require determining a threshold on the (cosine) similarity.
We also did not apply GPT-3.5 Turbo to the ranking task because of context size limitations and incurred costs for, e.g., running pair-wise ranking.

Our newly created dataset StatCodeSearch consists only of a single programming
language, which is R. There are other tools and programming languages that are used by researchers for statistical analysis. 
For a more extensive dataset in the domain of statistical research code, StatCodeSearch could be extended with code snippets from SPSS, STATA, SAS, and Python. 

Finally, it may seem counterintuitive that we have consulted GPT-3.5 Turbo for filtering, given its performance as a baseline for the matching is subpar. 
However, in the filtering step, the assessment of the pre-experiments showed that it was useful for discarding comments that were not suitable for the dataset, i.e., producing low numbers of false negatives. After this semi-automated pre-filtering, we manually filtered out the remaining code-comment pairs that were not suited for the dataset. 


\section*{Ethical Considerations}
Large language models come with ethical concerns regarding ethical bias, fairness, and transparency.
The purpose of the paper is to introduce a benchmark that aims at increasing our understanding of large language models' capabilities in the application of code search. Therefore, we do not see any new risks introduced by our paper.

Our dataset is derived solely from source code files released under public licenses enabling re-distribution. As described in \citet{husain2019codesearchnet} the CodeSearchNet dataset was filtered to include GitHub projects only with licenses explicitly permitting re-distribution.
The CoSQA and CodeSearchNet AdvTest sets are released under the Computational Use of Data Agreement (C-UDA) on the official CodeXGLUE repository\footnote{\url{https://github.com/microsoft/Computational-Use-of-Data-Agreement}}.
The newly created dataset StatCodeSearch consists of public content scraped from the Open Science Framework (OSF). All OSF content marked "Public" is available for commercial and non-commercial use according to the terms of use\footnote{\url{https://github.com/CenterForOpenScience/cos.io/blob/master/TERMS_OF_USE.md}}. Additionally we filtered out projects that did not include public licenses explicitly permitting modification and re-distribution. The source code files in the StatCodeSearch dataset include the public licenses \textit{Apache License, MIT License, CC-By Attribution 4.0, BSD 3-Clause, CC0 1.0 Universal, GNU General Public License, GNU Lesser General Public License}, and \textit{Mozilla Public License}.

\newpage
\bibliography{gencodesearchnet}

\begin{thebibliography}{34}
\expandafter\ifx\csname natexlab\endcsname\relax\def\natexlab#1{#1}\fi

\bibitem[{Allamanis et~al.(2015)Allamanis, Tarlow, Gordon, and
  Wei}]{allamanis2015bimodal}
Miltos Allamanis, Daniel Tarlow, Andrew Gordon, and Yi~Wei. 2015.
\newblock Bimodal modelling of source code and natural language.
\newblock In \emph{International conference on machine learning}, pages
  2123--2132. PMLR.

\bibitem[{Brown et~al.(2020)Brown, Mann, Ryder, Subbiah, Kaplan, Dhariwal,
  Neelakantan, Shyam, Sastry, Askell et~al.}]{brown2020language}
Tom Brown, Benjamin Mann, Nick Ryder, Melanie Subbiah, Jared~D Kaplan, Prafulla
  Dhariwal, Arvind Neelakantan, Pranav Shyam, Girish Sastry, Amanda Askell,
  et~al. 2020.
\newblock Language models are few-shot learners.
\newblock \emph{Advances in neural information processing systems},
  33:1877--1901.

\bibitem[{Cambronero et~al.(2019)Cambronero, Li, Kim, Sen, and
  Chandra}]{DBLP:conf/sigsoft/CambroneroLKS019}
Jos{\'{e}} Cambronero, Hongyu Li, Seohyun Kim, Koushik Sen, and Satish Chandra.
  2019.
\newblock \href {https://doi.org/10.1145/3338906.3340458} {When deep learning
  met code search}.
\newblock In \emph{Proceedings of the {ACM} Joint Meeting on European Software
  Engineering Conference and Symposium on the Foundations of Software
  Engineering, {ESEC/SIGSOFT} {FSE} 2019, Tallinn, Estonia, August 26-30,
  2019}, pages 964--974. {ACM}.

\bibitem[{Chen et~al.(2021)Chen, Tworek, Jun, Yuan, Pinto, Kaplan, Edwards,
  Burda, Joseph, Brockman et~al.}]{chen2021evaluating}
Mark Chen, Jerry Tworek, Heewoo Jun, Qiming Yuan, Henrique Ponde de~Oliveira
  Pinto, Jared Kaplan, Harri Edwards, Yuri Burda, Nicholas Joseph, Greg
  Brockman, et~al. 2021.
\newblock Evaluating large language models trained on code.
\newblock \emph{arXiv preprint arXiv:2107.03374}.

\bibitem[{Devlin et~al.(2019)Devlin, Ming-Wei, Kenton, and
  Toutanova}]{devlin2019bert}
Jacob Devlin, Chang Ming-Wei, Lee Kenton, and Kristina Toutanova. 2019.
\newblock {BERT}: Pre-training of deep bidirectional transformers for language
  understanding.
\newblock In \emph{Proceedings of NAACL-HLT}, pages 4171--4186.

\bibitem[{Feng et~al.(2020)Feng, Guo, Tang, Duan, Feng, Gong, Shou, Qin, Liu,
  Jiang et~al.}]{feng2020codebert}
Zhangyin Feng, Daya Guo, Duyu Tang, Nan Duan, Xiaocheng Feng, Ming Gong, Linjun
  Shou, Bing Qin, Ting Liu, Daxin Jiang, et~al. 2020.
\newblock Code{BERT}: A pre-trained model for programming and natural
  languages.
\newblock \emph{arXiv preprint arXiv:2002.08155}.

\bibitem[{Golchin and Surdeanu(2023)}]{golchin2023time}
Shahriar Golchin and Mihai Surdeanu. 2023.
\newblock Time travel in llms: Tracing data contamination in large language
  models.
\newblock \emph{arXiv preprint arXiv:2308.08493}.

\bibitem[{Goldenberg and Webb(2019)}]{goldenberg2019survey}
Igor Goldenberg and Geoffrey~I Webb. 2019.
\newblock Survey of distance measures for quantifying concept drift and shift
  in numeric data.
\newblock \emph{Knowledge and Information Systems}, 60(2):591--615.

\bibitem[{Goodfellow et~al.(2016)Goodfellow, Bengio, and
  Courville}]{goodfellowDeepLearning2016}
Ian~J. Goodfellow, Yoshua Bengio, and Aaron~C. Courville. 2016.
\newblock \emph{Deep {{Learning}}}.
\newblock Adaptive Computation and Machine Learning. {MIT Press}.

\bibitem[{Guo et~al.(2020)Guo, Ren, Lu, Feng, Tang, Liu, Zhou, Duan,
  Svyatkovskiy, Fu et~al.}]{guo2020graphcodebert}
Daya Guo, Shuo Ren, Shuai Lu, Zhangyin Feng, Duyu Tang, Shujie Liu, Long Zhou,
  Nan Duan, Alexey Svyatkovskiy, Shengyu Fu, et~al. 2020.
\newblock Graphcodebert: Pre-training code representations with data flow.
\newblock \emph{arXiv preprint arXiv:2009.08366}.

\bibitem[{Hindle et~al.(2012)Hindle, Barr, Su, Gabel, and
  Devanbu}]{naturalness_of_sw_hindle}
Abram Hindle, Earl~T. Barr, Zhendong Su, Mark Gabel, and Premkumar Devanbu.
  2012.
\newblock On the naturalness of software.
\newblock In \emph{Proceedings of the 34th International Conference on Software
  Engineering}, ICSE '12, page 837–847. IEEE Press.

\bibitem[{Hu et~al.(2023)Hu, Guo, Xie, Cordy, Papadakis, Ma, and
  Le~Traon}]{hu2023codes}
Qiang Hu, Yuejun Guo, Xiaofei Xie, Maxime Cordy, Mike Papadakis, Lei Ma, and
  Yves Le~Traon. 2023.
\newblock Codes: towards code model generalization under distribution shift.
\newblock In \emph{International Conference on Software Engineering (ICSE): New
  Ideas and Emerging Results (NIER)}.

\bibitem[{Huang et~al.(2021)Huang, Tang, Shou, Gong, Xu, Jiang, Zhou, and
  Duan}]{huang2021cosqa}
Junjie Huang, Duyu Tang, Linjun Shou, Ming Gong, Ke~Xu, Daxin Jiang, Ming Zhou,
  and Nan Duan. 2021.
\newblock Cosqa: 20,000+ web queries for code search and question answering.
\newblock In \emph{Proceedings of the 59th Annual Meeting of the Association
  for Computational Linguistics and the 11th International Joint Conference on
  Natural Language Processing (Volume 1: Long Papers)}, pages 5690--5700.

\bibitem[{Hupkes et~al.(2023)Hupkes, Giulianelli, Dankers, Artetxe, Elazar,
  Pimentel, Christodoulopoulos, Lasri, Saphra, Sinclair
  et~al.}]{hupkes2022state}
Dieuwke Hupkes, Mario Giulianelli, Verna Dankers, Mikel Artetxe, Yanai Elazar,
  Tiago Pimentel, Christos Christodoulopoulos, Karim Lasri, Naomi Saphra,
  Arabella Sinclair, et~al. 2023.
\newblock \href {https://doi.org/10.1038/s42256-023-00729-y} {A taxonomy and
  review of generalization research in {NLP}}.
\newblock \emph{Nature Machine Intelligence}, 5(10):1161--1174.

\bibitem[{Husain et~al.(2019)Husain, Wu, Gazit, Allamanis, and
  Brockschmidt}]{husain2019codesearchnet}
Hamel Husain, Ho-Hsiang Wu, Tiferet Gazit, Miltiadis Allamanis, and Marc
  Brockschmidt. 2019.
\newblock Codesearchnet challenge: Evaluating the state of semantic code
  search.
\newblock \emph{arXiv preprint arXiv:1909.09436}.

\bibitem[{Kabir et~al.(2023)Kabir, Udo-Imeh, Kou, and Zhang}]{kabir2023answers}
Samia Kabir, David~N Udo-Imeh, Bonan Kou, and Tianyi Zhang. 2023.
\newblock Who answers it better? an in-depth analysis of chatgpt and stack
  overflow answers to software engineering questions.
\newblock \emph{arXiv preprint arXiv:2308.02312}.

\bibitem[{Lhoest et~al.(2021)Lhoest, del Moral, Jernite, Thakur, von Platen,
  Patil, Chaumond, Drame, Plu, Tunstall et~al.}]{lhoest2021datasets}
Quentin Lhoest, Albert~Villanova del Moral, Yacine Jernite, Abhishek Thakur,
  Patrick von Platen, Suraj Patil, Julien Chaumond, Mariama Drame, Julien Plu,
  Lewis Tunstall, et~al. 2021.
\newblock Datasets: A community library for natural language processing.
\newblock In \emph{Proceedings of the 2021 Conference on Empirical Methods in
  Natural Language Processing: System Demonstrations}, pages 175--184.

\bibitem[{Li et~al.(2022{\natexlab{a}})Li, Gong, Shen, Qiu, Zhang, Yao, Qi,
  Jiang, Chen, and Duan}]{li-etal-2022-coderetriever}
Xiaonan Li, Yeyun Gong, Yelong Shen, Xipeng Qiu, Hang Zhang, Bolun Yao, Weizhen
  Qi, Daxin Jiang, Weizhu Chen, and Nan Duan. 2022{\natexlab{a}}.
\newblock \href {https://doi.org/10.18653/v1/2022.emnlp-main.187}
  {{C}ode{R}etriever: A large scale contrastive pre-training method for code
  search}.
\newblock In \emph{Proceedings of the 2022 Conference on Empirical Methods in
  Natural Language Processing}, pages 2898--2910, Abu Dhabi, United Arab
  Emirates. Association for Computational Linguistics.

\bibitem[{Li et~al.(2022{\natexlab{b}})Li, Gong, Shen, Qiu, Zhang, Yao, Qi,
  Jiang, Chen, and Duan}]{li2022coderetriever}
Xiaonan Li, Yeyun Gong, Yelong Shen, Xipeng Qiu, Hang Zhang, Bolun Yao, Weizhen
  Qi, Daxin Jiang, Weizhu Chen, and Nan Duan. 2022{\natexlab{b}}.
\newblock Coderetriever: Unimodal and bimodal contrastive learning.
\newblock \emph{arXiv preprint arXiv:2201.10866}.

\bibitem[{Lin et~al.(2022)Lin, Nogueira, and Yates}]{lin2022pretrained}
Jimmy Lin, Rodrigo Nogueira, and Andrew Yates. 2022.
\newblock \emph{Pretrained transformers for text ranking: Bert and beyond}.
\newblock Springer Nature.

\bibitem[{Liu et~al.(2019)Liu, Ott, Goyal, Du, Joshi, Chen, Levy, Lewis,
  Zettlemoyer, and Stoyanov}]{liu2019roberta}
Yinhan Liu, Myle Ott, Naman Goyal, Jingfei Du, Mandar Joshi, Danqi Chen, Omer
  Levy, Mike Lewis, Luke Zettlemoyer, and Veselin Stoyanov. 2019.
\newblock Ro{BERT}a: A robustly optimized {BERT} pretraining approach.
\newblock \emph{arXiv preprint arXiv:1907.11692}.

\bibitem[{Lu et~al.(2021)Lu, Guo, Ren, Huang, Svyatkovskiy, Blanco, Clement,
  Drain, Jiang, Tang et~al.}]{lu2021codexglue}
Shuai Lu, Daya Guo, Shuo Ren, Junjie Huang, Alexey Svyatkovskiy, Ambrosio
  Blanco, Colin Clement, Dawn Drain, Daxin Jiang, Duyu Tang, et~al. 2021.
\newblock Codexglue: A machine learning benchmark dataset for code
  understanding and generation.
\newblock In \emph{Thirty-fifth Conference on Neural Information Processing
  Systems Datasets and Benchmarks Track (Round 1)}.

\bibitem[{Luan et~al.(2019)Luan, Yang, Barnaby, Sen, and Chandra}]{LYB19}
Sifei Luan, Di~Yang, Celeste Barnaby, Koushik Sen, and Satish Chandra. 2019.
\newblock \href {https://doi.org/10.1145/3360578} {Aroma: code recommendation
  via structural code search}.
\newblock \emph{Proc. {ACM} Program. Lang.}, 3({OOPSLA}):152:1--152:28.

\bibitem[{Manning(2009)}]{manning2009introduction}
Christopher~D Manning. 2009.
\newblock \emph{An introduction to information retrieval}.
\newblock Cambridge university press.

\bibitem[{Mukherjee et~al.(2020)Mukherjee, Jermaine, and Chaudhuri}]{MJC20}
Rohan Mukherjee, Chris Jermaine, and Swarat Chaudhuri. 2020.
\newblock \href {https://doi.org/10.14778/3401960.3401972} {Searching a
  database of source codes using contextualized code search}.
\newblock \emph{Proc. {VLDB} Endow.}, 13(10):1765--1778.

\bibitem[{Neelakantan et~al.(2022)Neelakantan, Xu, Puri, Radford, Han, Tworek,
  Yuan, Tezak, Kim, Hallacy et~al.}]{neelakantan2022text}
Arvind Neelakantan, Tao Xu, Raul Puri, Alec Radford, Jesse~Michael Han, Jerry
  Tworek, Qiming Yuan, Nikolas Tezak, Jong~Wook Kim, Chris Hallacy, et~al.
  2022.
\newblock Text and code embeddings by contrastive pre-training.
\newblock \emph{arXiv preprint arXiv:2201.10005}.

\bibitem[{Niu et~al.(2023)Niu, Li, Ng, and Luo}]{niu2023crosscodebench}
Changan Niu, Chuanyi Li, Vincent Ng, and Bin Luo. 2023.
\newblock Crosscodebench: Benchmarking cross-task generalization of source code
  models.
\newblock \emph{arXiv preprint arXiv:2302.04030}.

\bibitem[{Ouyang et~al.(2022)Ouyang, Wu, Jiang, Almeida, Wainwright, Mishkin,
  Zhang, Agarwal, Slama, Ray et~al.}]{ouyang2022training}
Long Ouyang, Jeffrey Wu, Xu~Jiang, Diogo Almeida, Carroll Wainwright, Pamela
  Mishkin, Chong Zhang, Sandhini Agarwal, Katarina Slama, Alex Ray, et~al.
  2022.
\newblock Training language models to follow instructions with human feedback.
\newblock \emph{Advances in Neural Information Processing Systems},
  35:27730--27744.

\bibitem[{Sihler(2023)}]{sihler2023slicer}
Florian Sihler. 2023.
\newblock \href {https://doi.org/10.18725/OPARU-50107} {Constructing a static
  program slicer for~{R} programs}.
\newblock Master's thesis, Ulm University.

\bibitem[{Vinayakarao et~al.(2017)Vinayakarao, Sarma, Purandare, Jain, and
  Jain}]{VSP17}
Venkatesh Vinayakarao, Anita Sarma, Rahul Purandare, Shuktika Jain, and Saumya
  Jain. 2017.
\newblock \href {https://doi.org/10.1145/3018661.3018691} {{ANNE:} improving
  source code search using entity retrieval approach}.
\newblock In \emph{Proceedings of the Tenth {ACM} International Conference on
  Web Search and Data Mining, {WSDM} 2017, Cambridge, United Kingdom, February
  6-10, 2017}, pages 211--220. {ACM}.

\bibitem[{Wang et~al.(2023)Wang, Le, Gotmare, Bui, Li, and
  Hoi}]{wang2023codet5+}
Yue Wang, Hung Le, Akhilesh~Deepak Gotmare, Nghi~DQ Bui, Junnan Li, and
  Steven~CH Hoi. 2023.
\newblock Codet5+: Open code large language models for code understanding and
  generation.
\newblock \emph{arXiv preprint arXiv:2305.07922}.

\bibitem[{Wang et~al.(2021)Wang, Wang, Joty, and Hoi}]{wang2021codet5}
Yue Wang, Weishi Wang, Shafiq Joty, and Steven~CH Hoi. 2021.
\newblock Codet5: Identifier-aware unified pre-trained encoder-decoder models
  for code understanding and generation.
\newblock \emph{arXiv preprint arXiv:2109.00859}.

\bibitem[{Wolf et~al.(2019)Wolf, Debut, Sanh, Chaumond, Delangue, Moi, Cistac,
  Rault, Louf, Funtowicz et~al.}]{wolf2019huggingface}
Thomas Wolf, Lysandre Debut, Victor Sanh, Julien Chaumond, Clement Delangue,
  Anthony Moi, Pierric Cistac, Tim Rault, R{\'e}mi Louf, Morgan Funtowicz,
  et~al. 2019.
\newblock Huggingface's transformers: State-of-the-art natural language
  processing.
\newblock \emph{arXiv preprint arXiv:1910.03771}.

\bibitem[{Zhu et~al.(2022)Zhu, Jain, Suresh, Ravindran, Tipirneni, and
  Reddy}]{zhu2022xlcost}
Ming Zhu, Aneesh Jain, Karthik Suresh, Roshan Ravindran, Sindhu Tipirneni, and
  Chandan~K Reddy. 2022.
\newblock Xlcost: A benchmark dataset for cross-lingual code intelligence.
\newblock \emph{arXiv preprint arXiv:2206.08474}.

\end{thebibliography}
\bibliographystyle{acl_natbib}
\appendix
\section{Appendix}
\label{sec:appendix}
\subsection{Hyperparameter choice}
\label{sec:hyperparams}
For RoBERTa, we adopt best practices with a learning rate of $2 \times 10^{-5}$, $5$ epochs, batch size of $32$, weight decay of $0.01$, and sequence length of $512$.
For CodeBERT, we follow the original paper's parameters, while adjusting the sequence length to $512$ tokens for consistent comparisons. This involves a learning rate of $1 \times 10^{-5}$, a batch size of $32$, and $8$ epochs. 
Similarly, CodeT5+ adheres to its original parameters, but with a sequence length of $512$ tokens and a batch size of $32$ to ensure fairness. 

\begin{table}[htbp]
    \centering
    \small
    \caption{Hyperparameter settings for our language models}
    \label{tab:hyperparameters}
    \begin{tabular}{lccc}
        \toprule
        Hyperparameter & RoBERTa & CodeBERT & CodeT5+ \\
        \midrule
        Learning rate & 2 $\times 10^{-5}$ & 1 $\times 10^{-5}$ & 2 $\times 10^{-5}$ \\
        Epochs & 5 & 8 & 10 \\
        Batch size & 32 & 32 & 32 \\
        Weight decay & 0.01 & 0.01 & 0.01 \\
        Sequence length & 512 & 512 & 512 \\
        \bottomrule
    \end{tabular}
\end{table}

\subsection{Breakdown of CodeSearchNet Results}\label{app:breakdown}
Table~\ref{tab:codesearchnet} shows the breakdown of the CodeSearchNet by programming language. The numbers for in-distribution fine-tuning (fine-tuning for each individual language before testing) were taken from the original CodeBERT paper~\cite{feng2020codebert}, where for each sample 999 distractors were chosen. Although this makes the comparison to our numbers (with only 99 distractors) unfair, we can still observe a substantial decrease in performance when a model is fine-tuned on out-of-distribution data.

\begin{table*}[ht]
    \small
    \caption{Breakdown of the CodeSearchNet Results by Programming Language}
    \begin{tabular}{l|rr|rr|rr|rr|rr}
    \hline
    \textbf{Model} & \multicolumn{2}{c}{\textbf{Ruby}} & \multicolumn{2}{c}{\textbf{Go}} & \multicolumn{2}{c}{\textbf{PHP}} & \multicolumn{2}{c}{\textbf{Java}} & \multicolumn{2}{c}{\textbf{JavaScript}}\\
     & Acc & MRR & Acc & MRR & Acc & MRR & Acc & MRR & Acc & MRR \\
    \hline
    \textbf{Fine-tuned Models iid} & & & & & & & &\\
    RoBERTa \cite{feng2020codebert} & - & 0.6245 & - & 0.6809 & -  & 0.6576 & - & 0.6659  & - & 0.6060 \\
    CodeBERT \cite{feng2020codebert} & - & 0.6926 & - & 0.8400 & - & 0.7062 & - & 0.7484 & - & 0.7059\\
    \hline
    \textbf{Fine-tuned Models ood} & & & & & & & &\\
    RoBERTa & 92.71 & 0.1551 & 99.18 & 0.0469 & 89.73 & 0.0917 & 90.69 & 0.1228 & 90.85 & 0.1109\\
    CodeBERT & 89.71 & 0.1451 & 94.55 & 0.0668 & 87.21 & 0.0742 & 91.24 & 0.0822 & 90.10 & 0.0850\\
    CodeT5+ (encoder only) & 87.47 & 0.1003 & 91.86 & 0.0158 & 85.03 & 0.0568 & 86.04 & 0.0594 & 86.30 & 0.0756\\
    \hline
    \textbf{Zero-shot Models} & & & & & & & &\\
    CodeT5+ (encoder only) & - & 0.8398 & - & 0.8467 & - & 0.8000 & - & 0.7939 & - & 0.8185\\
    GPT 3.5 Turbo &58.70 & - &59.50 & - &56.41 & - &61.30 & - & 58.19 & -\\
    Ada 2 & - & 0.8942 & - & 0.9093& -& 0.8684& -& 0.8761 & -& 0.8782\\
    \hline
    \end{tabular}
    \label{tab:codesearchnet}
\end{table*}

\subsection{GPT 3.5 Turbo Prompts}
In the below sections, we report the prompts used for categorization and matching tasks for each of the test sets. GPT-3.5 was employed with default parameters, except for limit of 10 output tokens.

\subsubsection{Zero-shot code comment categorization}
\label{sec:code_categorization}
\vspace{0.5em}
\begin{minipage}{\linewidth}
\vspace{0.5em}
\noindent Task: Classify the code comment \{Input\} based on the categories provided below. If the comment doesn't fit into any of these categories, label it as `No Relevant Class'. Categories: [Statistical Test], [Statistical Modeling], [Data Variable], [Visualization]
\vspace{0.5em}
\end{minipage}
\vspace{0.5em}



\subsubsection{Matching}
\label{sec:prompts}
\subsubsection*{StatCodeSearch, CodeSearchNet and CodeSearchNet AdvTest}
\vspace{0.5em}
\begin{minipage}{\linewidth}
\vspace{0.5em}
\noindent Given a code comment and a \{Add the programming language name\} programming language code snippet, determine if the comment accurately represents the code's function. Respond with 'True' if the code matches the comment and 'False' if it does not. The input format is defined as “comment” “[CODESPLIT]” “code”. \{Input\}
\vspace{0.5em}
\end{minipage}
\vspace{0.5em}

\subsubsection*{CoSQA}
\vspace{0.5em}
\begin{minipage}{\linewidth}
\vspace{0.5em}
\noindent Given a search query and a Python programming language code snippet, determine if the query accurately represents the code's function. Respond with 'True' if the code matches the query and 'False' if it does not. The input format is defined as “query” “[CODESPLIT]” “code”. \{Input\}
\vspace{0.5em}
\end{minipage}
\vspace{0.5em}

\label{prompts}

\subsection{GenBench Evaluation Card}
In Table~\ref{tab:eval_card}, we provide the evaluation card proposed by \citep{hupkes2022state} for our experimental setups showcasing the different aspects of generalization our dataset studies.

\clearpage

\newcommand{\tabularwidth}{\textwidth}
\newcommand{\expone}{$\square$}
\newcommand{\exptwo}{$\bigtriangleup$}
\newcommand{\expthree}{$\bigcirc$}
        
\begin{table}[ht]
    \renewcommand{\arraystretch}{1.1}         
    \setlength{\tabcolsep}{0mm}         
    \begin{tabular}{|p{\tabularwidth}<{\centering}|}         
    \hline
    \rowcolor{gray!60}               
    \textbf{Motivation} \\               
    \footnotesize
    \begin{tabular}{p{0.25\tabularwidth}<{\centering} p{0.25\tabularwidth}<{\centering} p{0.25\tabularwidth}<{\centering} p{0.25\tabularwidth}<{\centering}}                 
    \textit{Practical} & \textit{Cognitive} & \textit{Intrinsic} & \textit{Fairness}\\
    \expone\hspace{0.8mm}\exptwo\hspace{0.8mm}\expthree\hspace{0.8mm}		
    & 		
    & 		
    & 		
    
    \vspace{2mm} \\
    \end{tabular}\\
                   
    \rowcolor{gray!60}               
    \textbf{Generalisation type} \\               
    \footnotesize
    \begin{tabular}{m{0.21\tabularwidth}<{\centering} m{0.2\tabularwidth}<{\centering} m{0.13\tabularwidth}<{\centering} m{0.13\tabularwidth}<{\centering} m{0.13\tabularwidth}<{\centering} m{0.2\tabularwidth}<{\centering}}                   
    \textit{Compositional} & \textit{Structural} & \textit{Cross Task} & \textit{Cross Language} & \textit{Cross Domain} & \textit{Robustness}\\
    & 		
    & 		
    & \expone\hspace{1.0mm}\exptwo\hspace{1.0mm}	
    & \exptwo\hspace{1.0mm}		
    & \hspace{10.0mm} \vspace{-2mm} \expthree	
    
    \vspace{2mm} \\
    \end{tabular}\\
                 
    \rowcolor{gray!60}             
    \textbf{Shift type} \\             
    \footnotesize
    \begin{tabular}{p{0.25\tabularwidth}<{\centering} p{0.25\tabularwidth}<{\centering} p{0.25\tabularwidth}<{\centering} p{0.25\tabularwidth}<{\centering}}                        
    \textit{Covariate} & \textit{Label} & \textit{Full} & \textit{Assumed}\\  
    \expone\hspace{0.8mm}\exptwo\hspace{0.8mm}\expthree\hspace{0.8mm}		
    & 		
    & 		
    & 		
    
    \vspace{2mm} \\
    \end{tabular}\\
                 
    \rowcolor{gray!60}             
    \textbf{Shift source} \\             
    \footnotesize
    \begin{tabular}{p{0.25\tabularwidth}<{\centering} p{0.25\tabularwidth}<{\centering} p{0.25\tabularwidth}<{\centering} p{0.25\tabularwidth}<{\centering}}                          
    \textit{Naturally occuring} & \textit{Partitioned natural} & \textit{Generated shift} & \textit{Fully generated}\\
    \expone\hspace{0.8mm}\exptwo\hspace{0.8mm}\expthree\hspace{0.8mm}		
    & 		
    & 		
    & 		
    
    \vspace{2mm} \\
    \end{tabular}\\
                 
    \rowcolor{gray!60}             
    \textbf{Shift locus}\\             
    \footnotesize
    \begin{tabular}{p{0.25\tabularwidth}<{\centering} p{0.25\tabularwidth}<{\centering} p{0.25\tabularwidth}<{\centering} p{0.25\tabularwidth}<{\centering}}                         
    \textit{Train--test} & \textit{Finetune train--test} & \textit{Pretrain--train} & \textit{Pretrain--test}\\
    & \expone\hspace{0.8mm}\exptwo\hspace{0.8mm}\expthree\hspace{0.8mm}		
    & 		
    & \hspace{10.0mm}\expone\hspace{0.8mm}\exptwo\hspace{0.8mm}\expthree\hspace{0.8mm}		
    
    \vspace{2mm} \\
    \end{tabular}\\
    
    \hline
    \end{tabular}
    \caption{GenBench Evaluation Card: \\
    \expone CodeSearchNet \\ \exptwo StatCodeSearch \\ \expthree CodeSearchNet Adv \& CoSQA}
    \label{tab:eval_card}
\end{table}

\end{document}